\documentclass[letterpaper]{article} 
\usepackage{aaai2026}  
\usepackage{times}  
\usepackage{helvet}  
\usepackage{courier}  
\usepackage[hyphens]{url}  
\usepackage{graphicx} 
\usepackage{natbib}  
\usepackage{caption} 
\usepackage{algorithm}
\usepackage{algorithmic}
\usepackage{multirow}
\usepackage{pifont}
\usepackage{amsmath}
\usepackage{bm}
\usepackage{amssymb}

\usepackage{makecell}
\usepackage{subfigure}
\usepackage{enumitem}
\usepackage{xcolor}
\usepackage{booktabs}
\usepackage{times}
\usepackage{helvet}
\usepackage{courier}
\usepackage{tabularx}

\newcommand{\ours}{APIE}
\newcommand{\ZSL}{ZSL}
\newcommand{\RSL}{RSL}
\newcommand{\other}{KD Sort}
\newcommand{\ActPrompt}{Active-Prompt}

\newcommand{\basex}[1]{\textcolor{gray!50}{\scriptsize \ $\uparrow${#1}}}
\newcommand{\basexx}[1]{\textcolor{gray!85}{\scriptsize \ $\uparrow${#1}}}
\newcommand{\basexd}[1]{\textcolor{gray!85}{\scriptsize \ $\downarrow${#1}}}

\usepackage{newfloat}
\usepackage{listings}
\DeclareCaptionStyle{ruled}{labelfont=normalfont,labelsep=colon,strut=off} 
\floatstyle{ruled}
\newfloat{listing}{tb}{lst}{}
\floatname{listing}{Listing}
\title{Reflect then Learn: Active Prompting for Information Extraction \\ Guided by Introspective Confusion}

\author{
    Dong Zhao\textsuperscript{\rm 1}\equalcontrib,
    Yadong Wang\textsuperscript{\rm 1}\equalcontrib, 
     Xiang Chen\textsuperscript{\rm 1}\footnotemark[2], 
     Chenxi Wang\textsuperscript{\rm 2}, 
     Hongliang Dai\textsuperscript{\rm 1},  \\
    Chuanxing Geng\textsuperscript{\rm 1}, 
    Shengzhong Zhang\textsuperscript{\rm 1}, 
    Shao-Yuan Li\textsuperscript{\rm 1}, 
    Sheng-Jun Huang\textsuperscript{\rm 1}\thanks{Corresponding authors.},
}
\affiliations{
    \textsuperscript{\rm 1}MIIT Key Laboratory of Pattern Analysis and Machine Intelligence, \\
    College of Computer Science and Technology,\\
    Nanjing University of Aeronautics and Astronautics \\
   \textsuperscript{\rm 2}Zhejiang University\\ 
  \{zhao\_dong, xiang\_chen\}@nuaa.edu.cn 
}

\begin{document}

\maketitle

\begin{abstract}
Large Language Models (LLMs) show remarkable potential for few-shot information extraction (IE), yet their performance is highly sensitive to the choice of in-context examples. Conventional selection strategies often fail to provide informative guidance, as they overlook a key source of model fallibility: confusion stemming not just from semantic content, but also from the generation of well-structured formats required by IE tasks. To address this, we introduce \textbf{A}ctive \textbf{P}rompting for \textbf{I}nformation \textbf{E}xtraction (\textbf{\ours}), a novel active prompting framework guided by a principle we term \textit{introspective confusion}. Our method empowers an LLM to assess its own confusion through a dual-component uncertainty metric that uniquely quantifies both Format Uncertainty (difficulty in generating correct syntax) and Content Uncertainty (inconsistency in extracted semantics). By ranking unlabeled data with this comprehensive score, our framework actively selects the most challenging and informative samples to serve as few-shot exemplars. Extensive experiments on four benchmarks show that our approach consistently outperforms strong baselines, yielding significant improvements in both extraction accuracy and robustness. Our work highlights the critical importance of a fine-grained, dual-level view of model uncertainty when it comes to building effective and reliable structured generation systems.
\end{abstract}

\begin{links}
    \link{Code}{https://github.com/NUAA-MMMI/APIE}
\end{links}

\section{Introduction}
Information Extraction (IE) serves as a foundational pillar for converting unstructured text into structured, machine-readable knowledge, thereby empowering a multitude of downstream applications such as knowledge graph construction, semantic search, and intelligent question answering~\cite{pai2024survey,ma2023large,mkgformer}. Traditionally, state-of-the-art IE systems have been dominated by fully supervised paradigms which necessitate extensive domain-specific labeled data for training. This reliance on large-scale annotation renders them costly and difficult to scale, particularly in specialized domains like medicine, law, or finance where expert annotation is a significant bottleneck and beyond.
\begin{figure}[t]
    \centering
    \includegraphics[width=1\linewidth]{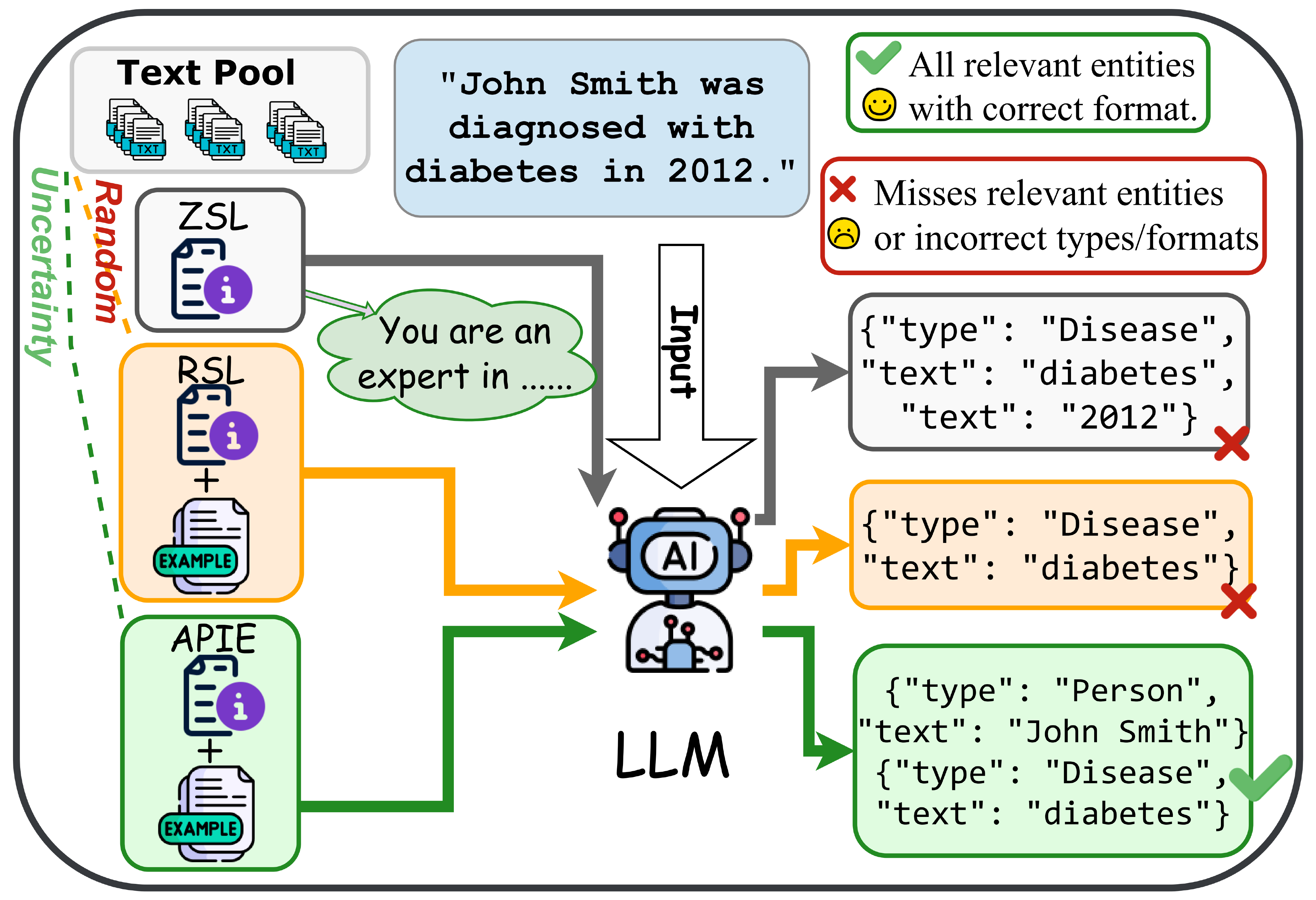}
    \caption{Illustration of Prompt Construction and Entity Extraction for ZSL, RSL, and \ours. Our approach achieves consistently higher extraction accuracy by selecting highquality examples via structured uncertainty scoring, while ZSL and RSL use fixed or random examples, leading to less accurate results across all domains.}
    \label{fig:fig1}
\end{figure}
The emergence of Large Language Models (LLMs) has catalyzed a paradigm shift toward few-shot learning, where models perform complex tasks by conditioning on a small number of in-context examples~\cite{brown2020language}. This prompt-based approach dramatically reduces the dependency on labeled data and offers unprecedented flexibility. However, the efficacy of LLMs in structured IE tasks is contingent on the quality of the provided exemplars. Prevailing selection strategies such as random sampling or simple semantic similarity often prove sub-optimal and lead to brittle performance as they fail to furnish the model with the most informative guidance for the intricate demands of IE and robustness~\cite{margatina2023active}.

This sensitivity stems from a fundamental -  yet often overlooked - dichotomy in the challenges inherent to structured generation tasks with LLMs. The model's fallibility arises not merely from semantic ambiguity in the source text, but also from the complex task of adhering to rigid output schemas . Existing uncertainty-guided prompting methods, while effective for classification tasks with discrete label spaces~\cite{diao2024active}, fall short in this context. They typically overlook a crucial dimension of model confusion: the struggle to maintain syntactic fidelity (i.e., structural correctness and parsability of the output) while simultaneously resolving semantic uncertainty (i.e., the correctness of the extracted content). This dual challenge of format and content necessitates a more nuanced approach to example selection.

To bridge this critical gap, we introduce \ours, an uncertainty-driven, training-free prompting framework designed for universal information extraction. Our approach is guided by a principle we term introspective confusion, which empowers the LLM to ``reflect" on its own generative process and identify samples that are most challenging and thus most valuable for learning. The model's internal confusion can be measured by analyzing the syntactic and semantic consistency across its own multiple, independently generated outputs. Specifically, we propose a dual-level introspective uncertainty metric that quantifies this confusion from two critical angles: 1) \textbf{Format-Level Uncertainty} which captures the model's struggle to produce structurally coherent and parsable outputs, measured through a combination of parsing failures and generation disagreement; and 2) \textbf{Content-Level Uncertainty} which assesses the semantic consistency of extracted information across multiple inferences using set-based divergence.

By ranking unlabeled data with this comprehensive uncertainty score, \ours~actively selects exemplars that are both structurally complex and semantically ambiguous. As illustrated in Figure~\ref{fig:fig1}, unlike zero-shot (ZSL) or random-sample (RSL) prompting which often lead to format errors or incorrect extractions, \ours~constructs highly informative prompts that guide the LLM toward producing accurate and well-structured outputs. We validate our framework through extensive experiments on four diverse IE benchmarks, demonstrating consistent and significant improvements over strong baselines across multiple LLMs.
Our main contributions are summarized as follows:
\begin{itemize}
    \item A novel active prompting framework, \textbf{\ours}, that uniquely introduces the principle of introspective confusion to better address the dual challenges of format and content generation in structured IE.
    \item The formulation of a dual-level introspective uncertainty estimation mechanism that uniquely combines format-level instability (parsing failures, generation variance) and content-level divergence (semantic inconsistency) to guide the selection of high-utility exemplars.
    \item Extensive empirical validation across four benchmark datasets and four LLMs, demonstrating that \ours~confidently establishes a new state-of-the-art for training-free universal information extraction, with significant gains in both overall accuracy and robustness.
\end{itemize}

\section{Related Work}

\subsection{Prompt-based Information Extraction with LLMs}

The advent of large language models has profoundly reshaped the landscape of information extraction (IE), shifting the focus from task-specific fine-tuning to versatile, prompt-based frameworks~\cite{knowprompt,liu2023pre}. These methods reframe IE as a language modeling problem, where a carefully crafted prompt guides an LLM to generate structured outputs for tasks like named entity recognition (NER), relation extraction (RE), and event extraction (EE)~\cite{DBLP:conf/emnlp/XieW0ZSYWZYWZWL22,uie}. This paradigm includes both zero-shot and few-shot prompting.

Despite their promise, the performance of these methods is highly sensitive to prompt design, especially the selection of in-context examples~\cite{diao2024active,han2023empirical}. Common strategies, such as random sampling or k-NN based on embedding similarity~\cite{wang2025gpt}, often fail to capture the nuanced structural and semantic complexities of IE tasks, resulting in unstable performance and a high variance across different example sets~\cite{liu2022makes}. To mitigate issues with output structure, some works have explored schema-guided prompting or post-processing parsers to enforce format constraints. However, these methods reactively correct or constrain the output, rather than proactively improving the model's intrinsic understanding of the task. They lack an adaptive selection mechanism that identifies exemplars capable of resolving the model's own confusion, a critical gap our work aims to fill.

\subsection{Uncertainty-Guided Prompt Optimization}

Uncertainty estimation is a cornerstone of active learning, where it is used to identify the most informative samples for labeling, thereby maximizing model performance with minimal annotation effort~\cite{DBLP:conf/sigir/LewisG94, gal2017deep,settles2009active}. Recently, these principles have been adapted to the LLM prompting paradigm. For instance, Active-Prompt~\cite{diao2024active} leverages uncertainty metrics like generation disagreement and entropy to select effective exemplars for complex reasoning tasks. Similarly, other studies have explored uncertainty to automate prompt engineering or select diverse demonstrations~\cite{margatina2023active, qian2024ape, DBLP:conf/aaai/ZhangYWL25}.

However, these pioneering efforts target classification or multi-choice reasoning tasks, where the output space is discrete and predefined. Their uncertainty metrics are ill-suited for the open-ended and highly structured nature of IE outputs~\cite{diao2024active}. The challenge in IE is not just choosing the correct label from a list, but instead involves generating a complex, composite structure with both correct content and valid syntax. Existing methods do not account for the model's uncertainty in generating a well-formed structure, a frequent point of failure for LLMs in IE.

In contrast, \ours~introduces a dual-level uncertainty framework tailored specifically for structured generation. While prior works measure disagreement at the final-answer level, our approach decomposes uncertainty into two orthogonal dimensions: syntactic fidelity and semantic uncertainty. This allows us to distinguish between a model that is confused about the content and one that is confused about the format. \ours~is the first framework to leverage such a holistic view of model confusion to actively and adaptively construct prompts for universal information extraction.

\begin{figure*}
    \centering
    \includegraphics[width=1\linewidth]{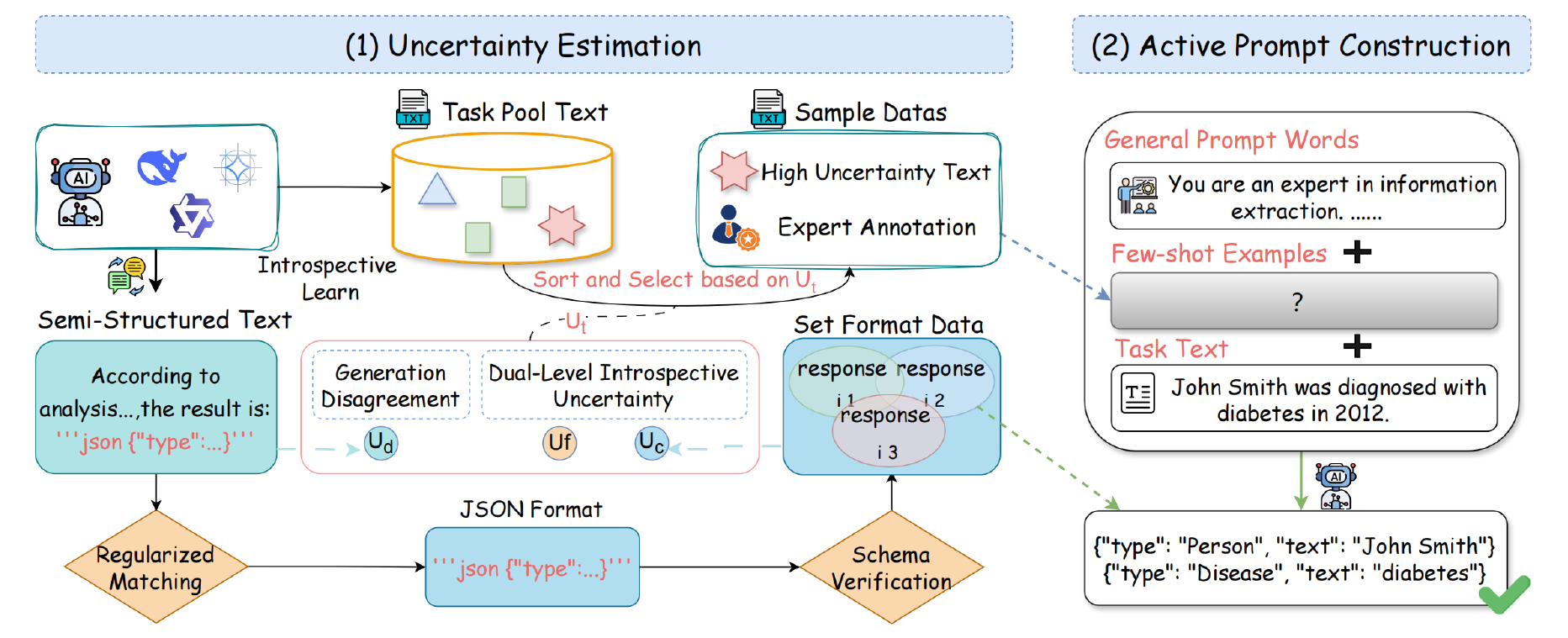}
    \caption{\ours~consists of two main stages. (1) Uncertainty Estimation: Given an unlabeled pool, candidate exemplars are evaluated using three complementary uncertainty signals—disagreement-level ($\mathcal{U}_d$), format-level ($\mathcal{U}_f$), and content-level ($\mathcal{U}_c$)—to compute a unified uncertainty score $\mathcal{U}_t$ and select high-utility examples. (2) Active Prompt Construction: The selected examples are incorporated into a structured prompt with task instructions and format guidance, enabling the language model to produce accurate, schema-conforming structured outputs (e.g., entity tuples). This framework enhances the informativeness and reliability of few-shot prompting without requiring labeled data.}
    \label{fig:fig2}
\end{figure*}

\section{Methodology}

Our proposed framework, \ours, introduces a novel active prompting strategy that selects high-utility exemplars by quantifying and leveraging the model's own introspective confusion. This section begins by formulating the task of universal information extraction as a prompt optimization problem under the few-shot learning setting. We then provide a comprehensive overview of the \ours~pipeline, followed by a detailed exposition of its core components: a dual-level uncertainty estimation mechanism and a structured, uncertainty-guided prompt construction process tailored for maximizing information gain.

\subsection{Problem Formulation}
We formulate information extraction (IE) as the task of learning a mapping function $\mathcal{F}: x \to y$, where $x$ is a piece of unstructured text from a domain $\mathcal{X}$, and $y$ is its corresponding structured representation (e.g., a set of entities, relations, or events) from a schema $\mathcal{Y}$. In a training-free, few-shot setting with an LLM $\mathcal{M}$, this mapping is conditioned on a prompt $P$ that contains task instructions and a set of $n$ exemplars $S = \{(x_j, y_j)\}_{j=1}^n$. The goal is to find an optimal set of exemplars $S^*$ from an unlabeled data pool $\mathcal{D}_u$ that maximizes the extraction performance on a test set:
\begin{equation}
S^* = \mathop{\arg\max}\limits_{S \subseteq \mathcal{D}_u, |S|=n} \mathbb{E}_{(x_i, y_i) \sim D_{test}} \left[ \mathcal{E}(\mathcal{M}(x_i, P(S)), y_i) \right],
\label{eq:selection}
\end{equation}
where $P(S)$ denotes the prompt constructed with exemplar set $S$, and $\mathcal{E}$ is a task-specific evaluation metric (e.g., F1-score). Directly solving this combinatorial optimization problem is intractable. Thus, \ours~proposes a principled proxy: selecting exemplars that maximize the model's estimated uncertainty as a surrogate objective.

\subsection{Output Schema and Task Setup}
\label{sec:schema}
To ensure consistency and enable automatic evaluation, we define a unified structured output schema strictly enforced for all IE tasks. The model's output must be a parsable JSON list of objects, where each object represents an extracted piece of information. For entities, objects must contain \texttt{"type"} and \texttt{"text"} keys; for relations, they require \texttt{"type"}, \texttt{"head"}, and \texttt{"tail"}. Our strict parser, $\mathcal{R}$, validates each generation against this schema; any deviation, such as malformed JSONs or missing required keys, is flagged as a format failure and contributes directly to the Format-Level Uncertainty ($\mathcal{U}_f$) calculation.

We also establish rigorous task boundaries for evaluation. Models are expected to produce an empty list (\texttt{[]}) for sentences containing no extractable information. For evaluation, we employ a strict matching protocol, where both the text span and type of an extracted element must exactly match the ground truth; partial matches are considered incorrect. This protocol is applied consistently across all methods for the final F1-score computation to ensure a fair and direct comparison without any exception whatsoever. 

\subsection{The Overview of our Framework}
The \ours~framework operationalizes the principle of Introspective Confusion through a systematic three-stage pipeline, illustrated in Figure~\ref{fig:fig2}. First, the framework probes an unlabeled pool by prompting the LLM to produce diverse outputs for each sample. These generations are then processed by our uncertainty metric, which integrates surface-level disagreement $\mathcal{U}_d$ with format-level $\mathcal{U}_f$ and content-level $\mathcal{U}_c$ scores into a unified measure of confusion. The pool is ranked accordingly, and the top high-uncertainty samples are selected as exemplars and placed into a pattern-constrained prompt with task instructions. This actively constructed prompt is then used to guide the LLM on new instances, yielding accurate and robust extractions.

\subsection{Generation Disagreement }
As a foundational baseline signal, we first measure the surface-level disagreement across $k$ generated outputs $\{\mathcal{M}^j_\theta(s_i)\}_{j=1}^k$ for a given sample $s_i$. We then quantify this using the average pairwise Levenshtein distance, which captures variations in the raw text output:
\begin{equation}
\mathcal{U}_d(s_i) = \frac{2}{k(k-1)} \sum_{1 \le j < l \le k} \text{Levenshtein}(\mathcal{M}^j_\theta(s_i), \mathcal{M}^l_\theta(s_i)),
\end{equation}
where a high $\mathcal{U}_d$ value signals a general state of overall output inconsistency without distinguishing between format and content discrepancies. In this sense, $\mathcal{U}_d$ can be viewed as a form of pre-format content uncertainty, capturing semantic variation prior to structural normalization.

\subsection{Dual-Level Introspective Uncertainty}
This is the core of our estimation strategy. Recognizing that failures in structured IE arise from distinct sources, we disentangle the model's confusion into two orthogonal dimensions. This decomposition allows us to move beyond monolithic uncertainty scores and gain a more granular understanding of why a model struggles with a particular sample, enabling a more targeted selection of exemplars.

\paragraph{Format-Level Uncertainty.} This metric quantifies the model's difficulty in adhering to the required output syntax. We define it as a combination of two signals:
\begin{itemize}
    \item \textbf{Parsing Failure Rate ($R_{fail}$):} The fraction of $k$ outputs that fail to be parsed by a strict structure parser $\mathcal{R}$. A high failure rate indicates the model is fundamentally struggling to generate syntactically valid structures. The parsing failure rate is calculated as follows:
\begin{equation}
R_{fail}(s_i) = \frac{1}{k} \sum_{j=1}^k \mathbb{I}[\mathcal{R}(\mathcal{M}^j_\theta(s_i)) = \text{FAIL}],
\end{equation}
where $\mathbb{I}$ is the indicator function.
    \item \textbf{Structural Disagreement:} For the successfully parsed outputs, we measure the variance in their structural composition (e.g., different sets of keys, varying list lengths). This captures more subtle inconsistencies in formatting.
\end{itemize}
The format-level uncertainty $\mathcal{U}_f$, a weighted sum of these components, is designed to directly measure the model's struggle with syntactic fidelity. 

\paragraph{Content-Level Uncertainty.} This metric assesses the semantic consistency of the information extracted in the successfully parsed outputs. Even if an LLM generates perfectly formatted outputs, it may still be uncertain about the actual content such as the boundaries of extracted spans or the correctness of predicted types and relations. For each sample $s_i$, we obtain a set of $k'$ valid structured outputs $B_i = \{b_1, b_2, \dots, b_{k'}\}$, where each $b_j$ is a set of extracted (type, text) tuples. We define content-level uncertainty as one minus the average pairwise Jaccard similarity, which measures the overlap of extracted information:
\begin{equation}
\mathcal{U}_c(s_i) = 1 - \frac{2}{k'(k'-1)} \sum_{1 \le j < l \le k'} \frac{|b_j \cap b_l|}{|b_j \cup b_l|}.
\end{equation}
A low Jaccard similarity (high $\mathcal{U}_c$) signifies that the model is generating semantically divergent extractions, indicating high uncertainty about the content. This signal is vital for identifying samples with inherent semantic ambiguity.

\subsection{Active Prompt Construction}
The final stage of our framework synthesizes the distinct uncertainty signals to construct a high-efficacy prompt. To ensure each signal contributes both fairly and appropriately to the final score, we first normalize the disagreement ($\mathcal{U}_d$), format ($\mathcal{U}_f$), and content ($\mathcal{U}_c$) uncertainties to a common [0, 1] scale. We then compute a unified uncertainty score, $\mathcal{U}_{total}$, for each candidate sample $s_i$ by taking a weighted combination of these normalized facets:
\begin{equation}
    \mathcal{U}_{total}(s_i) = \alpha \mathcal{U}_d(s_i) + \beta \mathcal{U}_f(s_i) + \gamma \mathcal{U}_c(s_i),
\end{equation}
where $\alpha, \beta$, and $\gamma$ are \textbf{key} hyperparameters that weigh the contribution of each uncertainty component. This design provides a high degree of flexibility, allowing for the prioritization of different uncertainty types depending on the specific task demands .

Samples from the pool $\mathcal{D}_u$ are ranked by their $\mathcal{U}_{total}$ score and the top-$n$ are selected. These selected texts, representing the most informative and challenging cases for the model, are then provided to a human expert for annotation to create the ground-truth labels. This active selection process significantly reduces the annotation burden as experts only need to label a small, targeted subset of data rather than a large random one, saving considerable human cost. The resulting high-quality exemplars are then seamlessly integrated into our prompt template to facilitate a more effective form of in-context learning, directly guiding the model toward more accurate and robust structured predictions.

\begin{table}[t]
\centering
\begin{tabular}{lccc}
\toprule
Dataset & $\left| Ent \right|$   & $\left| Rel \right|$ & \#Test \\
\midrule
ACE04   & 7     & --    & 812    \\
CoNLL03 & 4     & --    & 3,453  \\
CoNLL04 & 4     & 5     & 288    \\
SciERC  & 6     & 7     & 551    \\
\bottomrule
\end{tabular}
\caption{
  Detailed dataset statistics. Here, $\left | * \right |$ indicates the exact number of categories, and \# is the exact number of sentences in the specific subset.
}
\label{tab:details_datasets}
\end{table}

\begin{table*}[ht]
\centering
\scriptsize
\begin{tabular}{l|c|cc|cccc|c}
\toprule
\multirow{2}{*}{Models} 
& \multirow{2}{*}{Methods} 
& CoNLL03 
& ACE04-NER 
& \multicolumn{2}{c}{CoNLL04} 
& \multicolumn{2}{c}{SciERC} 
&\multirow{2}{*}{Average} 
\\
\cmidrule(lr){3-4} \cmidrule(lr){5-8}
& & NER F1-score↑ & NER F1-score↑ & NER F1-score↑ & RE F1-score↑ & NER F1-score↑ & RE F1-score↑ & F1-score↑ \\
\midrule
\multirow{5}{*}{\centering Gemma-3-12B} & \ZSL & 58.33\basex{0.00} & 22.48\basex{0.00} & 68.47\basex{0.00} & 20.59\basex{0.00} & 24.12\basex{0.00} & 14.53\basex{0.00} & 34.76\basex{0.00} \\
& \RSL & \textbf{62.96}\basexx{4.62} & 39.76\basexx{17.28} & 68.29\basexd{0.18} & 26.92\basexx{6.32} & \textbf{29.57}\basexx{5.45} & 14.43\basexd{0.11} & 40.32\basexx{5.56} \\
& \other & 59.80\basexx{1.47} & 40.11\basexx{17.63} & 68.58\basexx{0.11} & \textbf{36.71}\basexx{16.11} & 27.25\basexx{3.13} & 9.84\basexd{4.69} & 40.38\basexx{5.63} \\
& \ActPrompt & 58.91\basexx{0.57} & 43.61\basexx{21.13} & 65.76\basexd{2.71} & 28.38\basexx{7.78} & 25.45\basexx{1.33} & \textbf{19.25}\basexx{4.72} & 40.23\basexx{5.47} \\
& \ours & 60.07\basexx{1.73} & \textbf{44.42}\basexx{21.93} & \textbf{68.67}\basexx{0.20} & 28.82\basexx{8.23} & 26.21\basexx{2.09} & 16.85\basexx{2.32} & \textbf{40.84}\basexx{6.09} \\
\midrule
\multirow{5}{*}{\centering Qwen-2.5-14B} & \ZSL & 49.15\basex{0.00} & 10.78\basex{0.00} & 57.61\basex{0.00} & 13.69\basex{0.00} & 20.51\basex{0.00} & 18.08\basex{0.00} & 28.31\basex{0.00} \\
& \RSL & 63.35\basexx{14.20} & 30.05\basexx{19.27} & 63.35\basexx{5.74} & 30.33\basexx{16.64} & 27.94\basexx{7.43} & \textbf{19.23}\basexx{1.14} & 39.04\basexx{10.74} \\
& \other & 55.69\basexx{6.54} & 29.31\basexx{18.53} & 60.33\basexx{2.72} & 32.44\basexx{18.74} & 27.42\basexx{6.90} & 9.55\basexd{8.54} & 35.79\basexx{7.48} \\
& \ActPrompt & 64.29\basexx{15.15} & 30.03\basexx{19.25} & 63.31\basexx{5.70} & 29.61\basexx{15.91} & 29.11\basexx{8.60} & 13.74\basexd{4.34} & 38.35\basexx{10.04} \\
& \ours & \textbf{64.97}\basexx{15.82} & \textbf{32.61}\basexx{21.82} & \textbf{65.48}\basexx{7.87} & \textbf{32.57}\basexx{18.88} & \textbf{31.40}\basexx{10.89} & 14.11\basexd{3.97} & \textbf{40.19}\basexx{11.88} \\
\midrule
\multirow{5}{*}{\centering Deepseek-R1-14B} & \ZSL & 50.57\basex{0.00} & 17.76\basex{0.00} & 63.52\basex{0.00} & 27.30\basex{0.00} & 26.19\basex{0.00} & 11.06\basex{0.00} & 32.73\basex{0.00} \\
& \RSL & 57.88\basexx{7.30} & 31.50\basexx{13.73} & 63.97\basexx{0.46} & 27.83\basexx{0.54} & 26.87\basexx{0.68} & 11.30\basexx{0.24} & 36.56\basexx{3.83} \\
& \other & 51.47\basexx{0.90} & 31.24\basexx{13.47} & 63.48\basexd{0.04} & 33.11\basexx{5.81} & 29.17\basexx{2.98} & 10.75\basexd{0.31} & 36.53\basexx{3.80} \\
& \ActPrompt & 57.34\basexx{6.77} & 41.19\basexx{23.42} & 64.93\basexx{1.41} & 32.13\basexx{4.83} & 30.82\basexx{4.63} & 11.48\basexx{0.43} & 39.65\basexx{6.91} \\
& \ours & \textbf{59.28}\basexx{8.71} & \textbf{41.95}\basexx{24.18} & \textbf{66.32}\basexx{2.81} & \textbf{34.62}\basexx{7.33} & \textbf{33.09}\basexx{6.90} & \textbf{13.58}\basexx{2.52} & \textbf{41.47}\basexx{8.74} \\
\midrule
\multirow{5}{*}{\centering Deepseek-v3-660B} & \ZSL & 65.76\basex{0.00} & 24.87\basex{0.00} & 68.51\basex{0.00} & 31.74\basex{0.00} & 33.44\basex{0.00} & 20.30\basex{0.00} & 40.77\basex{0.00} \\
& \RSL & 69.71\basexx{3.95} & 33.31\basexx{8.44} & 71.29\basexx{2.79} & 42.32\basexx{10.58} & 40.43\basexx{6.99} & 20.35\basexx{0.04} & 46.23\basexx{5.46} \\
& \other & 69.25\basexx{3.49} & 38.62\basexx{13.75} & 72.57\basexx{4.06} & \textbf{51.01}\basexx{19.27} & 37.98\basexx{4.54} & 14.20\basexd{6.10} & 47.27\basexx{6.50} \\
& \ActPrompt & 72.48\basexx{6.72} & 47.33\basexx{22.46} & 72.75\basexx{4.25} & 47.66\basexx{15.92} & 41.89\basexx{8.46} & 18.90\basexd{1.41} & 50.17\basexx{9.40} \\
& \ours & \textbf{72.73}\basexx{6.97} & \textbf{48.56}\basexx{23.68} & \textbf{72.86}\basexx{4.36} & 50.20\basexx{18.46} & \textbf{42.05}\basexx{8.61} & \textbf{21.45}\basexx{1.15} & \textbf{51.31}\basexx{10.54} \\
\bottomrule
\end{tabular}
\caption{Overall performance (F1-score) comparison of \ours~against baseline methods across four information extraction benchmarks and four LLM backbones. We report F1-scores for both Named Entity Recognition (NER) and, where applicable, Relation Extraction (RE). The best-performing method for each metric is \textbf{bold}, and the absolute improvement over the \ZSL~baseline is reported alongside the main score. \ours~consistently reaches stable and highly competitive results.}
\label{tab:res_pope}
\end{table*}

\section{Experiments}

\subsection{Experimental Setup}

\paragraph{Tasks and Datasets.}
Our evaluation is conducted on four widely-used benchmarks spanning two core IE tasks: named entity recognition (NER) and relation extraction (RE). This diverse set, summarized in Table~\ref{tab:details_datasets}, includes both single-task (CoNLL03~\cite{tjongkimsang2003conll}, ACE04~\cite{ace2004-annotation}) and joint NER-RE settings (CoNLL04~\cite{roth-yih-2004-linear}, SciERC~\cite{luan-etal-2018-multi}), {which allows for a thorough assessment of framework generalizability. We adopt an end-to-end setting where models receive only raw text and must generate the target structures from it directly without any preprocessing.

\paragraph{Models.}
To evaluate performance across varying capabilities and architectures, we select four publicly available, open-source LLMs: \texttt{Gemma-3-12B}~\cite{gemmateam2025gemma3technicalreport}, \texttt{Qwen-2.5-14B}~\cite{qwen2025qwen25technicalreport}, \texttt{DeepSeek-R1-14B}~\cite{deepseekai2025deepseekr1incentivizingreasoningcapability}, and \texttt{DeepSeek-V3-660B}~\cite{deepseekai2025deepseekv3technicalreport}. This selection covers models with diverse architectures and scales, enabling a robust analysis of our method's scalability.

\paragraph{Baselines.}
We compare \ours~against four representative prompting strategies:

\begin{itemize}
    \item \textbf{Zero-Shot (ZSL):} The model performs extraction based only on the given task instructions, without the guidance of any in-context examples.
    \item \textbf{Random-Sample (RSL):} A standard few-shot baseline where exemplars are randomly selected.
    \item \textbf{KD Sort:} A curriculum-inspired strategy that selects exemplars based on knowledge density, defined as entities/relations per unit text length.

    \item \textbf{\ActPrompt:} An uncertainty-guided baseline that uses output disagreement on final answers to select exemplars, but lacks the structured uncertainty analysis.
\end{itemize}

\paragraph{Evaluation Metrics.}
Following standard practice, we use micro-averaged F1-score as the primary metric for all tasks. To ensure reproducibility, all reported results are averaged over multiple independent runs with different random seeds.

\subsection{Main Results}

\begin{figure}[t]
    \centering
    \includegraphics[width=0.48\textwidth]{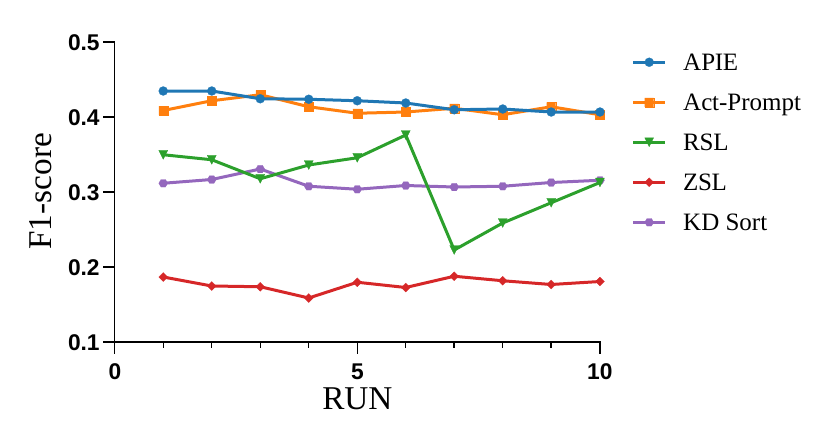}
    \caption{Performance stability comparison on the ACE04-NER dataset across 10 independent runs using the \texttt{DeepSeek-R1-14B} backbone.}
    \label{fig:stability_analysis}
\end{figure}

\paragraph{\textbf{For RQ1: Does \ours~consistently outperform established baselines in terms of extraction accuracy and robustness across different tasks and models?}} Our experiments demonstrate that \ours~consistently outperforms all baselines in both accuracy and robustness. Regarding accuracy, the main results in Table~\ref{tab:res_pope} show that our framework achieves the highest average F1-score across nearly all backbones. The performance gains are particularly pronounced on complex joint-extraction tasks. For instance, with the \texttt{DeepSeek-R1-14B} model on the SciERC dataset, \ours~achieves a 13.58 F1-score in relation extraction, marking a 18.3\% relative improvement over the Active-Prompt baseline (11.48). This highlights the superior ability of our structured uncertainty metrics to identify informative exemplars for complex schema. Beyond superior accuracy, \ours~also delivers greater robustness. As visualized in Figure~\ref{fig:stability_analysis}, the performance of the Random-Sample (RSL) baseline fluctuates dramatically across runs, underscoring its unreliability. In contrast, \ours~produces consistent results with minimal variance, confirming that our approach constructs stable and generalizable prompts.

\begin{figure}[t]
    \centering
    \includegraphics[width=0.48\textwidth]{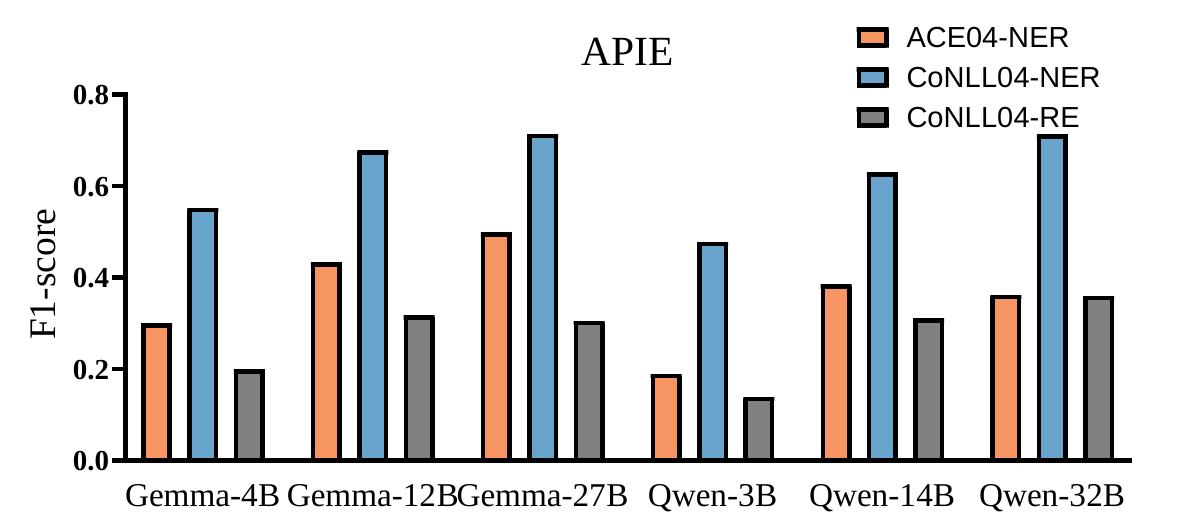}
    \caption{F1-score comparison across different model scales using \ours~on three different datasets.Here, Gemma refers to \texttt{Gemma-3}, and Qwen refers to \texttt{Qwen-2.5}.}
    \label{fig:modelscale}
\end{figure}

\paragraph{\textbf{For RQ2: How does the effectiveness of \ours~scale with the capacity of the backbone large language models?}} We investigate the impact of model scale on performance with results shown in Figure~\ref{fig:modelscale}. Our analysis reveals that \ours's advantage is most pronounced on smaller-scale models such as \texttt{Gemma-3-4B} and \texttt{Qwen-2.5-3B}, where performance gaps against strong baselines exceed 10 F1 points on tasks such as ACE04-NER. This suggests that our structured uncertainty-guided prompting compensates effectively for the limited capacity and reasoning ability of compact LLMs. As model size increases, the overall performance of all prompting strategies improves, and inter-method performance gaps gradually narrow, indicating that large-capacity models are inherently more resilient to suboptimal exemplar selection. Nonetheless, \ours~maintains a consistent and measurable lead even at the largest scale (e.g., \texttt{Deepseek-R1-14B}), affirming the broad utility of our approach. Importantly, we observe that scaling alone does not resolve performance instability: as shown in Figure~\ref{fig:stability_analysis}, random prompting continues to yield high variance even on high-capacity models, whereas \ours~achieves remarkably stable results across all scales, underscoring its robustness and practicality in real-world deployments.

\paragraph{\textbf{For RQ3: How are the different uncertainty signals distributed, and what does their behavior reveal about the nature of model confusion in IE tasks?}} We analyze the empirical distributions of the uncertainty scores from our experiments. We visualize these distributions using violin plots in Figure~\ref{fig:Analysis of Uncertainty}. The uncertainty signals are complementary and capture orthogonal aspects of difficulty in model behavior. The violin plots reveal that the three signals have distinct distributional characteristics. For instance, we observe cases where a sample has minimal format uncertainty ($\mathcal{U}_f=0.0$) but maximal content uncertainty ($\mathcal{U}_c=1.0$), indicating the model can generate the correct syntax but is confused about the semantic content. Conversely, other samples show high format uncertainty but low content uncertainty.

\begin{figure}[t]
    \centering
    \includegraphics[width=0.48\textwidth]{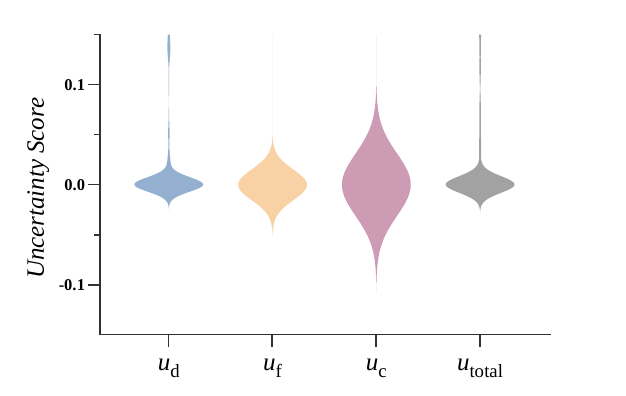}
    \caption{Uncertainty score distributions for $\mathcal{U}_d$, $\mathcal{U}_f$, and $\mathcal{U}_c$ on the ACE04-NER dataset, evaluated using the \texttt{DeepSeek-R1-14B} model.}
    \label{fig:Analysis of Uncertainty}
\end{figure}

\begin{table*}[t]
\centering
\small 

\begin{tabular}{l cc cc cc cc}
\toprule
\textbf{Methods} & \multicolumn{2}{c}{\textbf{ACE04-NER}} & \multicolumn{2}{c}{\textbf{CoNLL04 NER}} & \multicolumn{2}{c}{\textbf{CoNLL04 RE}} & \multicolumn{2}{c}{\textbf{Average}} \\
\cmidrule(lr){2-3} \cmidrule(lr){4-5} \cmidrule(lr){6-7} \cmidrule(lr){8-9}
& Acc. & F1 & Acc. & F1 & Acc. & F1 & Acc. & F1 \\
\midrule
Full \ours & \textbf{35.04} & \textbf{41.95} & \textbf{55.71} & \textbf{66.32} & \textbf{32.46} & \textbf{34.62} & \textbf{41.07} & \textbf{47.63} \\
w/o Pattern Prompt (p) & 25.38 & 32.58 & 49.44 & 59.77 & 10.04 & 10.66 & 28.29 & 34.34 \\
w/o Format Uncertainty ($\mathcal{U}_f$) & 31.71 & 39.39 & 53.55 & 64.17 & 29.10 & 30.88 & 38.12 & 44.81 \\
w/o Content Uncertainty ($\mathcal{U}_c$) & 33.49 & 41.04 & 53.93 & 64.41 & 31.23 & 32.90 & 39.55 & 46.12 \\
w/o Disagreement ($\mathcal{U}_d$) & 26.56 & 32.19 & 52.61 & 62.86 & 30.13 & 31.98 & 36.43 & 42.34 \\
\bottomrule
\end{tabular}
\caption{
Ablation study on the effectiveness of \ours~components across NER and RE tasks. Removing any component leads to performance degradation, with $\mathcal{U}_d$ and pattern prompts being particularly critical.
}
\label{tab:ablation_f1}
\end{table*}

\begin{figure}[t]
\centering
\footnotesize
\begin{tabular}{@{}p{\linewidth}@{}}
\toprule
\textbf{Input Sentence:} \\
\textit{On the 22nd, the sixteen NATO countries asked the armed forces of the Serbian people of Bosnia-Herzegovina to withdraw...} \\
\midrule
\textbf{Ground Truth Entities:} \\
\{ \{`type': `Organization', `text': `the sixteen NATO countries'\}, \{`type': `Location', `text': `Bosnia-Herzegovina'\}, \{`type': `Location', `text': `Gerlaridy city'\}, \{`type': `Group', `text': `Serbian'\}, \{`type': `Organization', `text': `NATO'\}, \{`type': `Organization', `text': `United Nations'\} \} \\
\midrule
\textbf{Gemma-3-12B:} 
\{ \{`type': `Organization', `text': `NATO'\}, \{`type': `Organization', `text': `United Nations'\}, \{`type': `Location', `text': `Bosnia-Herzegovina'\}, \{`type': `Organization', `text': `the Serbian army'\} \} \\[0.2em]
\textbf{Qwen-2.5-14B:} 
\{ \{`type': `Organization', `text': `the sixteen NATO countries'\}, \{`type': `Organization', `text': `the armed forces of the Serbian people of Bosnia-Herzegovina'\} \} \\[0.2em]
\textbf{Deepseek-R1-14B:} 
\{ \{`type': `Organization', `text': `sixteen NATO countries'\}, \{`type': `Location', `text': `Bosnia-Herzegovina'\}, \{`type': `Organization', `text': `United Nations humanitarian aid'\} \} \\
\midrule
\textbf{Uncertainty:} $\mathcal{U}_d=0.828$, $\mathcal{U}_f=0.0$, $\mathcal{U}_c=0.556$ \\
\midrule
\textbf{Explanation:} 
Despite consistent format ($\mathcal{U}_f=0.0$), models diverge in semantic interpretation—some view ‘sixteen NATO countries’ as a geopolitical entity, others as an organization. 
\ours{} detects this content-level ambiguity ($\mathcal{U}_c=0.556$) and prioritizes such samples for targeted refinement. \\
\bottomrule
\end{tabular}
\caption{Case study illustrating content-level uncertainty despite consistent format understanding.}
\label{fig:case-study}
\end{figure}

\begin{figure}[t]
    \centering
    \includegraphics[width=0.48\textwidth]{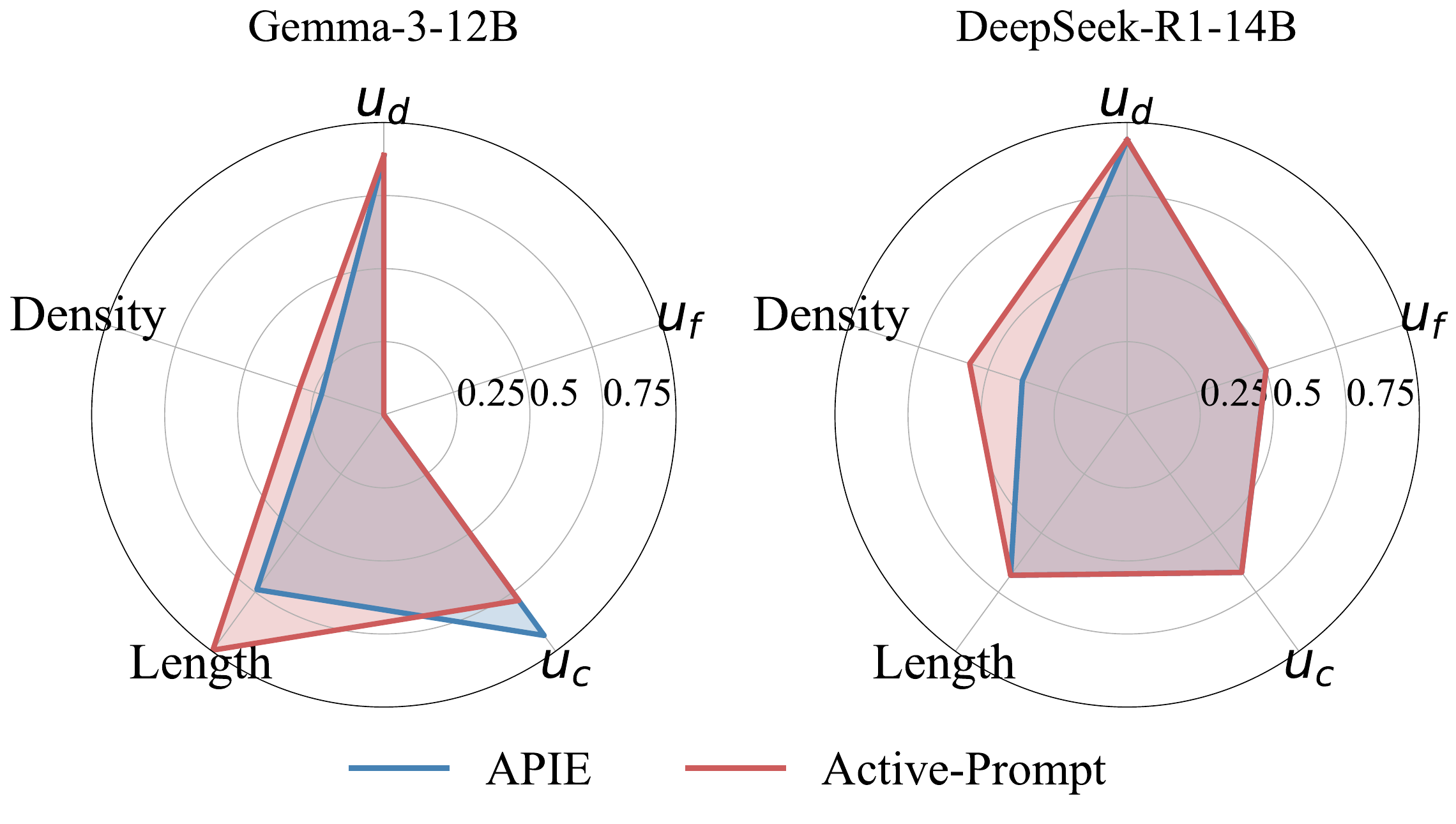}
    \caption{Comparison of exemplar characteristics selected by \ours~and Active-Prompt on the ACE04-NER dataset.}
    \label{fig:casestudy}
\end{figure}

\subsection{Analysis}
\paragraph{Ablation Study.}To dissect the contribution of each component in our framework, we conduct an ablation study using the \texttt{DeepSeek-R1-14B} model. As shown in Table~\ref{tab:ablation_f1}, removing any single component—disagreement-based uncertainty ($\mathcal{U}_d$), format-level uncertainty ($\mathcal{U}_f$), content-level uncertainty ($\mathcal{U}_c$), or the pattern-guided prompt—leads to a noticeable performance drop across both NER and RE tasks. This indicates that each module contributes non-trivially to the overall performance of \ours. Notably, the exclusion of $\mathcal{U}_d$ causes the largest decline in F1-score, underscoring its critical role as a foundational signal of model uncertainty. The $\mathcal{U}_f$ and $\mathcal{U}_c$ metrics further complement this base signal, refining the selection toward structurally and semantically ambiguous samples. Additionally, the removal of pattern prompts yields the sharpest degradation in RE accuracy, suggesting that structure-aware prompt design plays an essential role in guiding the model toward consistent and schema-compliant outputs. These results validate the necessity of each module in constructing informative and stable prompts for reliable few-shot IE.

\paragraph{Case Study.}To characterize the high-utility exemplars selected by \ours, we perform individual cases (Figure~\ref{fig:case-study}) and a qualitative analysis combining both aggregate trends (Figure~\ref{fig:casestudy}). The analysis reveals a key distinction in selection strategies between our method and baselines. As shown in the radar plot, the Active-Prompt baseline, which relies on surface-level disagreement, tends to select for samples that are simply longer, denser, and cause high raw disagreement ($\mathcal{U}_d$). In contrast, \ours~demonstrates a more targeted selection strategy. While not always choosing the longest or densest samples, our framework shows a distinct and significant advantage in identifying instances with high content uncertainty ($\mathcal{U}_c$) and format uncertainty ($\mathcal{U}_f$). This suggests that our dual-level metric successfully moves beyond simple textual variance to pinpoint samples that are truly challenging in their semantic interpretation and structural requirements. The input sentence is syntactically straightforward, leading to zero format uncertainty ($\mathcal{U}_f = 0.0$). However, it elicits significant semantic confusion among different models regarding entity boundaries , resulting in high content uncertainty ($\mathcal{U}_c=0.556$). By identifying such targeted, semantically ambiguous challenges , \ours~enables more effective teaching for nuanced cases.

\section{Conclusion and Outlook}

In this work, we introduce \ours, a novel active prompting framework guided by introspective confusion to improve exemplar selection for few-shot information extraction. By disentangling model confusion into format and content uncertainty, \ours~selects challenging exemplars to build more informative and robust prompts. Experiments show consistent gains over strong baselines, highlighting the value of structured uncertainty for improving LLM reliability. Future directions include more efficient uncertainty estimation, automatic hyperparameter tuning, and extensions to structured generation tasks like code generation, potentially combined with new diversity-based metrics.

\section*{Acknowledgments}
 
This work was supported by the National Natural Science Foundation of China (Nos. 62506166, U2441285, 62222605),  the Natural Science Foundation
of Jiangsu Province (No. BK20251365), the China Postdoctoral Science Foundation (No. 2025M774283), the Yongjiang Talent Introduction Programme (No. 2021A-156-G) and
the Ningbo Natural Science Foundation (No. 2024J020).
This research is also sponsored by the DiDi GAIA Collaborative Research Funds (No. CCF-DiDi GAIA202507) and CAAI-MindSpore Open Fund (CAAIXSJLJJ 2025 MindSpore 01), developed on OpenI Community.

\bibliography{aaai2026}

\makeatletter
\@ifundefined{isChecklistMainFile}{
  \newif\ifreproStandalone
  \reproStandalonetrue
}{
  \newif\ifreproStandalone
  \reproStandalonefalse
}
\makeatother

\setlength{\pdfpagewidth}{8.5in}
\setlength{\pdfpageheight}{11in}
\frenchspacing

\setlength{\leftmargini}{20pt}
\makeatletter
\def\@listi{\leftmargin\leftmargini \topsep .5em \parsep .5em \itemsep .5em}
\def\@listii{\leftmargin\leftmarginii \labelwidth\leftmarginii \advance\labelwidth-\labelsep \topsep .4em \parsep .4em \itemsep .4em}
\def\@listiii{\leftmargin\leftmarginiii \labelwidth\leftmarginiii \advance\labelwidth-\labelsep \topsep .4em \parsep .4em \itemsep .4em}
\makeatother

\setcounter{secnumdepth}{0}
\renewcommand\thesubsection{\arabic{subsection}}
\renewcommand\labelenumi{\thesubsection.\arabic{enumi}}

\newcounter{checksubsection}
\newcounter{checkitem}[checksubsection]

\newcommand{\checksubsection}[1]{%
  \refstepcounter{checksubsection}%
  \paragraph{\arabic{checksubsection}. #1}%
  \setcounter{checkitem}{0}%
}

\newcommand{\checkitem}{%
  \refstepcounter{checkitem}%
  \item[\arabic{checksubsection}.\arabic{checkitem}.]%
}
\newcommand{\question}[2]{\normalcolor\checkitem #1 #2 \color{blue}}
\newcommand{\ifyespoints}[1]{\makebox[0pt][l]{\hspace{-15pt}\normalcolor #1}}

\ifreproStandalone
\begin{document}
\fi

\section*{Reproducibility Checklist}


\checksubsection{General Paper Structure}
\begin{itemize}

\question{Includes a conceptual outline and/or pseudocode description of AI methods introduced}{(yes/partial/no/NA)}
yes

\question{Clearly delineates statements that are opinions, hypothesis, and speculation from objective facts and results}{(yes/no)}
yes

\question{Provides well-marked pedagogical references for less-familiar readers to gain background necessary to replicate the paper}{(yes/no)}
yes

\end{itemize}
\checksubsection{Theoretical Contributions}
\begin{itemize}

\question{Does this paper make theoretical contributions?}{(yes/no)}
no

	\ifyespoints{\vspace{1.2em}If yes, please address the following points:}
        \begin{itemize}
	
	\question{All assumptions and restrictions are stated clearly and formally}{(yes/partial/no)}
	NA

	\question{All novel claims are stated formally (e.g., in theorem statements)}{(yes/partial/no)}
	NA

	\question{Proofs of all novel claims are included}{(yes/partial/no)}
	NA

	\question{Proof sketches or intuitions are given for complex and/or novel results}{(yes/partial/no)}
	NA

	\question{Appropriate citations to theoretical tools used are given}{(yes/partial/no)}
	NA

	\question{All theoretical claims are demonstrated empirically to hold}{(yes/partial/no/NA)}
	NA

	\question{All experimental code used to eliminate or disprove claims is included}{(yes/no/NA)}
	NA
	
	\end{itemize}
\end{itemize}

\checksubsection{Dataset Usage}
\begin{itemize}

\question{Does this paper rely on one or more datasets?}{(yes/no)}
yes

\ifyespoints{If yes, please address the following points:}
\begin{itemize}

	\question{A motivation is given for why the experiments are conducted on the selected datasets}{(yes/partial/no/NA)}
	yes

	\question{All novel datasets introduced in this paper are included in a data appendix}{(yes/partial/no/NA)}
	NA

	\question{All novel datasets introduced in this paper will be made publicly available upon publication of the paper with a license that allows free usage for research purposes}{(yes/partial/no/NA)}
	NA

	\question{All datasets drawn from the existing literature (potentially including authors' own previously published work) are accompanied by appropriate citations}{(yes/no/NA)}
	yes

	\question{All datasets drawn from the existing literature (potentially including authors' own previously published work) are publicly available}{(yes/partial/no/NA)}
	yes

	\question{All datasets that are not publicly available are described in detail, with explanation why publicly available alternatives are not scientifically satisficing}{(yes/partial/no/NA)}
	NA

\end{itemize}
\end{itemize}

\checksubsection{Computational Experiments}
\begin{itemize}

\question{Does this paper include computational experiments?}{(yes/no)}
yes

\ifyespoints{If yes, please address the following points:}
\begin{itemize}

	\question{This paper states the number and range of values tried per (hyper-) parameter during development of the paper, along with the criterion used for selecting the final parameter setting}{(yes/partial/no/NA)}
	partial

	\question{Any code required for pre-processing data is included in the appendix}{(yes/partial/no)}
	yes

	\question{All source code required for conducting and analyzing the experiments is included in a code appendix}{(yes/partial/no)}
	yes

	\question{All source code required for conducting and analyzing the experiments will be made publicly available upon publication of the paper with a license that allows free usage for research purposes}{(yes/partial/no)}
	yes
        
	\question{All source code implementing new methods have comments detailing the implementation, with references to the paper where each step comes from}{(yes/partial/no)}
	yes

	\question{If an algorithm depends on randomness, then the method used for setting seeds is described in a way sufficient to allow replication of results}{(yes/partial/no/NA)}
	yes

	\question{This paper specifies the computing infrastructure used for running experiments (hardware and software), including GPU/CPU models; amount of memory; operating system; names and versions of relevant software libraries and frameworks}{(yes/partial/no)}
	yes

	\question{This paper formally describes evaluation metrics used and explains the motivation for choosing these metrics}{(yes/partial/no)}
	yes

	\question{This paper states the number of algorithm runs used to compute each reported result}{(yes/no)}
	yes

	\question{Analysis of experiments goes beyond single-dimensional summaries of performance (e.g., average; median) to include measures of variation, confidence, or other distributional information}{(yes/no)}
	yes

	\question{The significance of any improvement or decrease in performance is judged using appropriate statistical tests (e.g., Wilcoxon signed-rank)}{(yes/partial/no)}
	partial

	\question{This paper lists all final (hyper-)parameters used for each model/algorithm in the paper’s experiments}{(yes/partial/no/NA)}
	partial

\end{itemize}
\end{itemize}
\ifreproStandalone
\end{document}


\maketitle

\section{Appendix A : Implementation Details}
This section provides a detailed description of the experimental setup to ensure full reproducibility of our results.

\subsection{A.1 Model Configuration}
All models used in the experiments are free and open-source large language models. Except for \texttt{DeepSeek-V3-660B}, the models were downloaded from Ollama and run locally. Additionally, identical models can be found on the Hugging Face Hub. The \texttt{DeepSeek-V3-660B} model utilizes the API calling platform publicly provided by DeepSeek. The model specifications referenced in our experiments are listed in Table~\ref{tab:model_details}.

\begin{table}[hbt!]
\centering
\begin{tabular}{ll}
\toprule
\textbf{Model Name} & \textbf{Reference / Platform} \\
\midrule
Gemma-3-4B/12B/27B & \makecell[l]{ollama.com/library/gemma3}\\
Qwen-2.5-3B/14B/32B & \makecell[l]{ollama.com/library/qwen2.5}\\
DeepSeek-R1-14B & \makecell[l]{ollama.com/library/deepseek-r1}\\
DeepSeek-V3-660B & \makecell[l]{api.deepseek.com}\\
\bottomrule
\end{tabular}
\caption{Backbone models used in the experiments.}
\label{tab:model_details}
\end{table}

All experiments were performed on a server equipped with an NVIDIA L40S GPU. The framework was implemented using Python 3.10 and the Ollama library (v0.9.3).

\subsection{A.2 Hyperparameters}
The primary hyperparameters used for the experiments in the main text are detailed in Table~\ref{tab:parameters}. For JSON parsing, we used the \texttt{jsonschema} (v4.23.0) library to ensure strict validation. The maximum token length for generation was set to the default value for each respective model.

\begin{table}[hbt!]
\centering

\begin{tabular}{@{}lc@{}} 
\toprule
\textbf{Parameter} & \textbf{Value} \\
\midrule
\multicolumn{2}{l}{\textit{Uncertainty Estimation Stage}} \\
\quad $k$ (Number of Probing Outputs) & 3 \\
\quad Temperature & 0.8 \\
\quad Number of Exemplars & 2 \\
\midrule
\multicolumn{2}{l}{\textit{Final Inference Stage}} \\
\quad $\alpha$ (Disagreement Weight) & 0.8 \\
\quad $\beta$ (Format Uncertainty Weight) & 0.1 \\
\quad $\gamma$ (Content Uncertainty Weight) & 0.1 \\
\quad Temperature & 0.8 \\
\quad Number of Exemplars & 3 \\
\bottomrule
\end{tabular}
\caption{Primary hyperparameters used in the \ours~framework.}
\label{tab:parameters}
\end{table}

\begin{table}[hbt!]
\centering
\scriptsize
\begin{tabular}{lcccc}
\toprule
\multirow{2}{*}{Models} 
& \multirow{2}{*}{k}
& ACE04-NER
& \multicolumn{2}{c}{CoNLL04} \\
\cmidrule(lr){3-3} \cmidrule(lr){4-5}
& &NER F1-score↑ & NER F1-score↑ & RE F1-score↑\\
\midrule
\multirow{3}{*}{\centering Gemma-3-12B} 
& 2 & 41.83 & 70.38 & 25.71 \\
& 3 & 44.42 & 68.67 & 28.82 \\
& 5 & 43.66 & 70.20 & 16.63 \\
\midrule
\multirow{3}{*}{\centering Qwen-2.5-14B} 
& 2 & 35.60 & 68.23 & 28.80 \\
& 3 & 32.61 & 65.48 & 32.57 \\
& 5 & 40.73 & 65.98 & 32.76 \\
\bottomrule
\end{tabular}
\caption{Effect of varying the number of probing samples ($k$) on F1-score. The optimal value of $k$ can depend on the model and dataset, with $k=3$ offering a strong balance of performance and efficiency.}
\label{tab:result_k}
\end{table}

\begin{table*}[hbt!]
\centering
\scriptsize
\begin{tabular}{l|c|cc|cccc|c}
\toprule
\multirow{2}{*}{Models} 
& {Parameter} 
& CoNLL03 
& ACE04-NER 
& \multicolumn{2}{c}{CoNLL04} 
& \multicolumn{2}{c}{SciERC}
& \multirow{2}{*}{Average} \\
\cmidrule(lr){3-4} \cmidrule(lr){5-8}
& ($\alpha$ , $\beta$, $\gamma$) & NER F1-score↑ & NER F1-score↑ & NER F1-score↑ & RE F1-score↑ & NER F1-score↑ & RE F1-score↑ & F1-score↑ \\
\midrule
\multirow{3}{*}{\centering Gemma-3-12B} 
& (0.33, 0.33, 0.33) & 60.02 & 45.25 & 67.68 & 20.64 & 28.30 & 14.68 & 39.46 \\
& (0.3, 0.5, 0.2) & 59.85 & 44.84 & 68.99 & 24.77 & 26.97 & 17.71 & 40.52 \\
& (0.5, 0.2, 0.3) & 61.49 & 34.32 & 67.92 & 23.60 & 25.19 & 17.78 & 38.38 \\
\midrule
\multirow{3}{*}{\centering Qwen-2.5-14B} 
& (0.33, 0.33, 0.33) & 65.06 & 29.43 & 63.59 & 28.74 & 32.63 & 14.32 & 38.96 \\
& (0.3, 0.5, 0.2) & 65.08 & 29.39 & 63.15 & 32.73 & 31.12 & 14.69 & 39.36 \\
& (0.5, 0.2, 0.3) & 64.70 & 32.35 & 65.23 & 31.64 & 31.46 & 14.15 & 39.92 \\
\midrule
\multirow{3}{*}{\centering Deepseek-R1-14B} 
& (0.33, 0.33, 0.33) & 61.94 & 40.96 & 68.01 & 39.34 & 35.36 & 12.19 & 42.96 \\
& (0.3, 0.5, 0.2) & 59.91 & 45.02 & 67.89 & 35.56 & 35.00 & 11.18 & 42.43 \\
& (0.5, 0.2, 0.3) & 58.70 & 43.34 & 67.91 & 36.23 & 35.17 & 10.95 & 42.05 \\
\midrule
\multirow{3}{*}{\centering Deepseek-v3-660B} 
& (0.33, 0.33, 0.33) & 69.73 & 49.28 & 72.83 & 36.86 & 44.23 & 20.39 & 48.89 \\
& (0.3, 0.5, 0.2) & 70.73 & 49.63 & 73.14 & 39.58 & 41.50 & 19.89 & 49.08 \\
& (0.5, 0.2, 0.3) & 72.48 & 48.93 & 72.16 & 50.06 & 41.99 & 21.95 & 51.26 \\
\bottomrule
\end{tabular}
\caption{Performance (F1-score) of \ours~under different uncertainty weight configurations ($\alpha, \beta, \gamma$). The results show that the framework is robust to different weighting schemes, though prioritizing disagreement or content signals can be beneficial depending on the model and task.}
\label{tab:result_Parameter}
\end{table*}

\section{Appendix B : Additional Experimental Analysis}
\label{sec:appendix_additional_exp}

\subsection{B.1 Hyperparameter Sensitivity}
To assess the robustness of \ours~to its internal hyperparameters, we conducted a sensitivity analysis on the uncertainty weights $\alpha$, $\beta$, and $\gamma$. As shown in Table~\ref{tab:result_Parameter}, we tested several weighting schemes. The results indicate that while performance is relatively stable across different configurations, a higher weight on disagreement ($\alpha=0.8$ in the main text, or $\alpha=0.5$ here) generally yields strong results. Notably, for the most capable model, DeepSeek-V3-660B, prioritizing content-related signals (e.g., $\alpha=0.5, \gamma=0.3$) can lead to state-of-the-art performance in complex RE tasks, highlighting the flexibility of our framework.


\subsection{B.2 Effect of Probing Sample Count ($k$)}
The number of generated outputs, $k$, is a critical parameter that balances uncertainty estimation quality and computational cost. We varied $k$ from 2 to 5 and measured its effect on F1-score, as detailed in Table~\ref{tab:result_k}. The results show a non-monotonic relationship; performance does not strictly increase with $k$. For instance, with the Gemma-3-12B model, performance on ACE04-NER peaked at $k=3$. This suggests that a small number of probing samples is often sufficient to capture a reliable uncertainty signal, and a very high $k$ may not always yield further benefits. Our choice of $k=3$ in the main experiments represents an effective trade-off between accuracy and efficiency.


\section{Appendix C : Full Prompt Templates and Exemplars}
\label{sec:appendix_prompt}
To ensure clarity and reproducibility, this section details the general-purpose prompt template used in our experiments. Our prompt design is modular and consists of several key components, each serving a specific purpose to guide the LLM effectively. This multi-part structure is designed to provide the model with a clear, unambiguous, and highly informative context, thereby maximizing its ability to perform accurate and reliable structured extraction.

The prompt begins with a Task Definition, which outlines the objective and assigns a role to the model (e.g., ``You are an expert in information extraction''). This is followed by explicit Schema and Format Instructions, where we define the target entity and relation types and specify the required output format (a JSON list of objects). This instruction is critical for ensuring structural consistency and serves as the ground truth for calculating Format-Level Uncertainty. The core of the prompt is the Exemplars section. Unlike standard few-shot prompting, this section is populated with high-utility examples actively selected by our \ours~framework for their high Introspective Confusion scores. Finally, the prompt concludes with the Target Input, which is the new, unseen sentence for extraction. Figure~\ref{fig:appendixeTemplates} provides a concrete instantiation of this template.

\begin{figure}[hbt!]
    \centering
    \includegraphics[width=1.0\linewidth]{figures/prompt.pdf} 
    \caption{An illustration of the prompt structure provided to the LLM. It includes a task definition, schema instructions, one or more high-utility exemplars (input-output pairs), and the final target question.}
    \label{fig:appendixeTemplates}
\end{figure}